\documentclass[journal]{IEEEtran}

\usepackage{cite}
\usepackage{amsmath,amssymb,amsfonts}
\usepackage{algorithmic}
\usepackage{graphicx}
\usepackage{textcomp}
\usepackage{xcolor}
\usepackage{hyperref}
\usepackage{url}
\usepackage{booktabs}

\hypersetup{
    colorlinks=true,
    linkcolor=blue,
    urlcolor=blue,
    citecolor=blue
}

\renewcommand{\footnoterule}{%
    \kern -3pt
    \hrule width \columnwidth height 0.4pt
    \kern 2pt
}

\def\BibTeX{{\rm B\kern-.05em{\sc i\kern-.025em b}\kern-.08em
    T\kern-.1667em\lower.7ex\hbox{E}\kern-.125emX}}

\begin{document}

\title{HighSync: High-Quality Lip Synchronization via Latent Diffusion Models}

\author{Saeid~Firouzi~Daghigh,
        Majid~Iranpour~Mobarakeh,
        Mostafa~Alavi,
        and~Mahdi~Bagheri%
\thanks{\textit{(Corresponding author: Saeid Firouzi Daghigh.)}}%
\thanks{This work has been submitted to the IEEE for possible publication. 
Copyright may be transferred without notice, after which this version 
may no longer be accessible.}
\thanks{S. Firouzi Daghigh and M. Iranpour Mobarekeh are with the Department of Computer Engineering and Information Technology, Payame Noor University, Tehran, Iran (e-mail: saeedmr881@gmail.com; iranpour@pnu.ac.ir).}%
\thanks{M. Alavi is an independent researcher (e-mail: mostafa.alavi25@gmail.com).}%
\thanks{M. Bagheri is an independent researcher (e-mail: mahdid.m.2000@gmail.com).}}

\maketitle

\begin{abstract}
We present HighSync, an end-to-end diffusion-based framework for high-fidelity lip synchronization that generates photorealistic talking-face videos aligned with arbitrary input audio. Existing approaches consistently struggle to reconcile image quality with synchronization accuracy, producing either visually degraded outputs or temporally inconsistent lip movements. HighSync addresses both challenges simultaneously and, to our knowledge, is the first lip sync model to operate natively at $512{\times}512$ resolution, positioning it as a viable solution for professional production environments such as the film and broadcast industries. Central to our approach is the identification and systematic elimination of a data leakage phenomenon that has silently undermined temporal modeling in prior work, preventing models from developing a genuine dependence on the audio signal. Comprehensive evaluations across both perceptual quality and synchronization accuracy metrics confirm that HighSync achieves state-of-the-art performance on both fronts. Source code, pre-trained models, and supplementary video results are publicly available at: \url{https://github.com/saeed5959/high_sync}
\end{abstract}

\begin{IEEEkeywords}
lip synchronization, talking-face generation, audio-visual alignment
\end{IEEEkeywords}

\section{Introduction}

Lip synchronization is the task of modifying the lip region and lower facial area of a talking-face video to align with a target speech signal, while preserving the subject's identity, head pose, and overall visual quality~\cite{prajwal2020lip, mukhopadhyay2024diff2lip, yu2024make, zhang2024musetalk}. The task has broad practical relevance, spanning multilingual film dubbing, post-production video editing, virtual avatar creation, and educational content localization-contexts~\cite{prajwal2020lip, yu2024make} in which reshooting or re-recording is either impractical or prohibitively costly~\cite{mukhopadhyay2024diff2lip}.

A persistent requirement across all these applications is photorealistic output: generated frames must be visually indistinguishable from authentic footage, free of blurring, boundary artifacts, and dental distortions. Despite steady progress in the field, the overwhelming majority of existing methods fall well short of this requirement. Most operate at resolutions of $96{\times}96$, $128{\times}128$, or at best $256{\times}256$ pixels~\cite{prajwal2020lip, mukhopadhyay2024diff2lip, yu2024make}, and produce outputs that suffer from pronounced visual degradation in the lip and teeth regions, rendering them unsuitable for deployment in high-fidelity production pipelines.

Generative Adversarial Network (GAN)-based methods including Wav2Lip~\cite{prajwal2020lip}, StyleSync~\cite{guan2023stylesync}, StyleLipSync~\cite{ki2023stylelipsync}, and VideoReTalking~\cite{cheng2022videoretlaking} have long constituted the dominant paradigm in lip sync research. However, GAN architectures are fundamentally constrained by training instability and mode collapse~\cite{che2016mode}, which limits their ability to scale to diverse, high-resolution datasets and to generalize across the wide variability of in-the-wild faces. Diffusion models, by contrast, have demonstrated superior generative capacity across a broad spectrum of image and video synthesis tasks, particularly for high-resolution face generation~\cite{rombach2022high}. The first application of diffusion models to lip sync, Diff2Lip~\cite{mukhopadhyay2024diff2lip}, demonstrated improved perceptual quality but remained limited in resolution and synchronization accuracy. LatentSync~\cite{li2024latentsync} subsequently proposed a more principled latent diffusion framework and introduced comprehensive empirical studies on SyncNet convergence, achieving notable improvements in synchronization; nevertheless, its temporal consistency and overall sync accuracy remain areas for improvement.

A core difficulty in lip sync, less frequently acknowledged, is that synchronization is inherently a temporal property: it cannot be measured from any single frame in isolation, but only across a contiguous sequence of frames exhibiting coherent lip motion~\cite{prajwal2020lip, li2024latentsync}. This constraint gives rise to two competing modeling strategies. The first relies on external models, most commonly SyncNet~\cite{chung2016out}, to impose synchronization constraints on independently generated frames during training~\cite{prajwal2020lip, mukhopadhyay2024diff2lip, li2024latentsync}. The second, more architecturally integrated approach builds temporal dependencies directly into the model via a motion module~\cite{guo2023animatediff}, temporal self-attention layers, and optimizes synchronization within the model itself. While the first strategy has been widely adopted, it is contingent on reliable SyncNet convergence, which is notoriously difficult to achieve and sensitive to numerous hyperparameter and preprocessing choices~\cite{li2024latentsync}. The second approach is theoretically more principled, but has seen limited success: LatentSync~\cite{li2024latentsync} reports a failed attempt to use a motion module without identifying its cause.

In this work, we pinpoint the precise reason for this failure. Through systematic experimentation, we demonstrate that the underlying cause is a \textit{data leakage problem} with two distinct sources: (1) frame-level variation in face bounding box height induced by per-frame face detection, and (2) the biomechanical correlation between upper facial muscle dynamics and lip movements, both of which are amplified when the model processes consecutive frames jointly via a motion module. This leakage allows the model to reconstruct the original lip trajectory from contextual upper-face cues, entirely bypassing its dependence on the audio signal. Upon identifying and eliminating both sources, we successfully integrate a motion module into our training pipeline, enabling the model to learn rich temporal lip dynamics across batches of 12 consecutive frames, without any reliance on SyncNet as a training supervisor.

Built on Stable Diffusion 1.5~\cite{rombach2022high} as its generative backbone, HighSync incorporates a dedicated Reference U-Net for fine-grained identity preservation and cross-attention-based audio conditioning via Whisper~\cite{radford2023robust} embeddings. The resulting system achieves state-of-the-art synchronization and visual quality at $512{\times}512$ resolution through a clean, two-stage training procedure free of adversarial supervision.

Our primary contributions are as follows:
\begin{itemize}
    \item We propose HighSync, the first end-to-end lip synchronization model to generate temporally coherent, high-fidelity videos at $512{\times}512$ resolution without relying on SyncNet supervision.
    \item We systematically identify and resolve a data leakage problem that undermines temporal audio conditioning in motion-module-based lip sync models, and propose concrete preprocessing and architectural fixes that enable effective motion module training.
\end{itemize}

\section{Related Work}

\subsection{Non-Diffusion-Based Lip Synchronization}

The lip sync literature has been dominated by GAN-based approaches for several years. Wav2Lip~\cite{prajwal2020lip} established the foundational paradigm of using a frozen, pretrained SyncNet~\cite{chung2016out} discriminator to supervise a lip sync generator, demonstrating that an accurate external synchronization signal is essential for producing convincing lip motion and reconstruction losses alone are insufficient. StyleSync~\cite{guan2023stylesync} retained this supervisory scheme while adopting StyleGAN2 as its generator backbone, yielding improved visual fidelity at the cost of increased computational complexity. VideoReTalking~\cite{cheng2022videoretlaking} decomposes the lip sync process into three specialized stages, face reenactment, synchronization, and identity-aware refinement, achieving better results on high-resolution inputs but at the expense of a brittle multi-stage pipeline. DINet~\cite{zhang2023dinet} introduces a deformation-based inpainting network that warps feature maps conditioned on driving audio, circumventing the need for explicit landmark estimation while producing competitive visual quality. MuseTalk~\cite{zhang2024musetalk} is of particular relevance: it adopts the U-Net backbone of Stable Diffusion for latent-space inpainting but replaces the diffusion process with adversarial training, effectively functioning as an efficient, one-step GAN operating in latent space. Its two-stage training procedure, comprising a Facial Abstract Pretraining stage followed by Lip-Sync Adversarial Finetuning with Informative Frame Sampling and Dynamic Margin Sampling, achieves real-time inference at $256{\times}256$ resolution, though synchronization accuracy remains bounded by GAN training dynamics. Despite the computational efficiency of GAN-based methods, they are fundamentally limited in their ability to scale to large, diverse datasets~\cite{che2016mode}, and consistently produce blurry, artifact-laden outputs that degrade perceptual quality.

\subsection{Diffusion-Based Lip Synchronization}

The emergence of latent diffusion models~\cite{rombach2022high} has opened new possibilities for high-quality lip synchronization. Diff2Lip~\cite{mukhopadhyay2024diff2lip} was the first method to pose lip sync as an audio-conditioned inpainting problem within a pixel-space diffusion framework, introducing both perceptual (LPIPS) and sequential adversarial losses to encourage visual quality and inter-frame consistency. While it surpassed GAN-based methods in perceptual image quality, its resolution remained low and its synchronization accuracy was limited. LatentSync~\cite{li2024latentsync} marked a substantial step forward by formulating lip sync as a fully end-to-end latent diffusion problem without any intermediate motion representation. Its primary technical contributions include an extensive empirical analysis of SyncNet convergence, identifying batch size, number of input frames, and data preprocessing (particularly affine transformation and audio-visual offset adjustment) as critical factors, and the Temporal REPresentation Alignment (TREPA) method, which uses VideoMAE-v2~\cite{wang2023videomae} temporal representations to enforce frame-sequence consistency as a soft auxiliary loss, without adding parameters. LatentSync achieves 94\% SyncNet accuracy on HDTF and outperforms prior methods across multiple metrics, but its frame-by-frame generation approach and reliance on SyncNet leave room for improvement in temporal coherence and synchronization robustness. EchoMimic~\cite{chen2024echomimic}, while primarily targeting audio-driven portrait animation rather than lip sync, is directly relevant to our architectural choices: it introduces a Reference U-Net for identity preservation, a Temporal-Attention module for inter-frame coherence, and an overlapped-context inference strategy in which the last two motion frames of one generation round are reused as context for the next, a mechanism we adopt and extend in our framework.

\section{Methodology}

\subsection{Model Architecture}

The HighSync framework is built upon Stable Diffusion 1.5 (SD 1.5)~\cite{rombach2022high}, extended to operate on sequences of 12 consecutive frames through the integration of a temporal motion module~\cite{chen2024echomimic}. Two conditioning signals drive the generation process: a reference image, which supplies visual identity and facial texture information to compensate for the masked lip region in the input frames; and a driving audio signal, which encodes the target speech content to determine lip shape and movement.

\begin{figure*}[t]
    \centering
    \includegraphics[width=\textwidth]{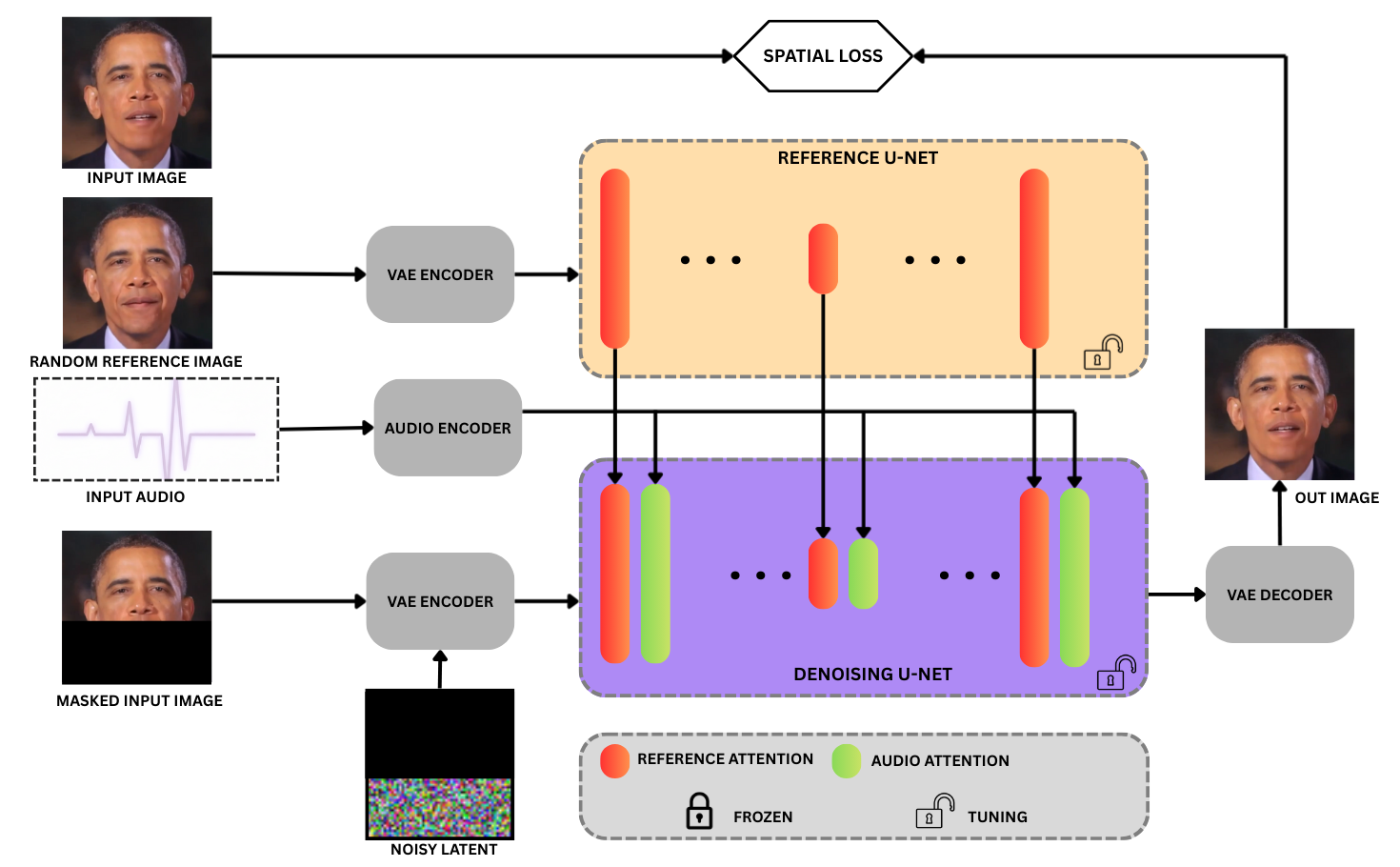}
    \caption{Overview of the HighSync Stage 1 training framework. The model processes a single masked input frame alongside a randomly selected reference image from the same video. Both are encoded into the latent space via a shared VAE Encoder. The Reference U-Net, whose architecture mirrors that of SD 1.5, processes the reference image and injects fine-grained identity features into every transformer block of the Denoising U-Net via Reference-Attention layers. Audio features extracted by the Whisper encoder are integrated through Audio-Attention cross-attention layers. The Denoising U-Net, initialized from SD 1.5, predicts the denoised latent, which is decoded by the VAE Decoder to produce the output frame. A Spatial Loss is computed between the decoded output and the ground-truth frame in pixel space. In this stage, the Reference U-Net and the Denoising U-Net is set to tuning mode while the remaining components, excluding the motion module (no motion module is present in this stage), are trained end-to-end. No motion module is present in Stage 1.}
    \label{fig:stage1}
\end{figure*}

\begin{figure*}[t]
    \centering
    \includegraphics[width=\textwidth]{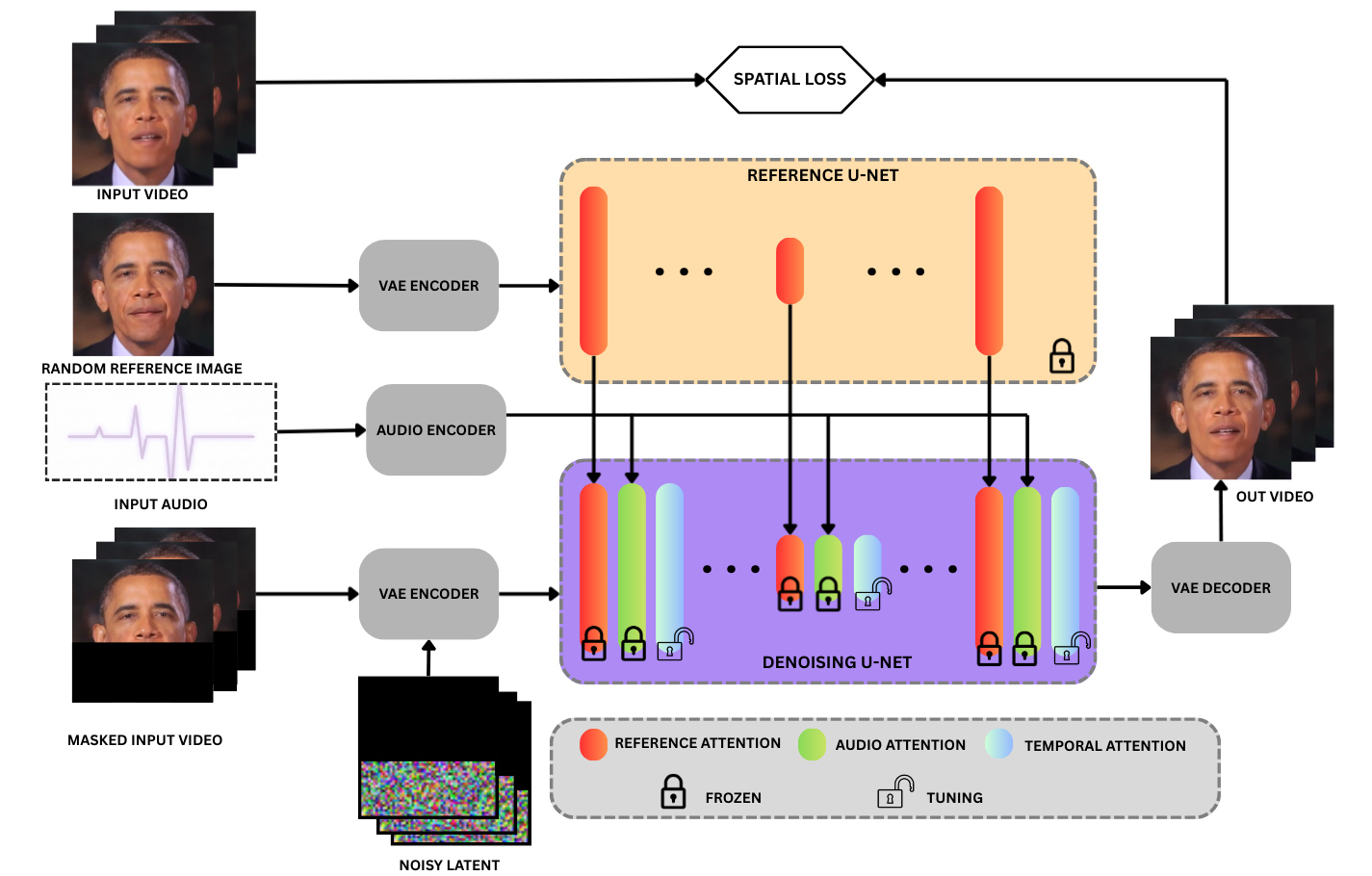}
    \caption{Overview of the HighSync Stage 2 training framework. All components from Stage 1, the Reference U-Net, the audio encoder, and the Denoising U-Net, are frozen. Only the Temporal-Attention (motion module) layers, newly inserted into the Denoising U-Net, are trained. The model now receives sequences of 12 consecutive masked input frames, 12 corresponding noisy latents, and a single randomly selected reference image shared across all frames in the batch. Temporal-Attention layers model dependencies across the frame sequence to produce smooth, temporally coherent lip motion. The Spatial Loss is computed over all 12 decoded output frames. During inference, each of the 12 frames is provided with its own ground-truth reference image, rather than a single shared one, substantially reducing per-frame artifacts.}
    \label{fig:stage2}
\end{figure*}

\subsubsection{Reference Image Conditioning}

We explored two straightforward conditioning strategies before arriving at our final design. Channel-wise concatenation of the reference image with the masked input frames, as used in several prior works~\cite{prajwal2020lip, mukhopadhyay2024diff2lip, li2024latentsync}, failed to provide sufficient identity signal, particularly under large pose or expression variation. Image embedding via a pretrained encoder captured high-level semantic content but lacked the fine-grained texture detail required for faithful lip and teeth reconstruction.

We therefore utilize a dedicated Reference U-Net, similar to EchoMimic~\cite{chen2024echomimic}, that mirrors the full encoder-decoder architecture of the Denoising U-Net and operates in parallel with it. At each transformer block, the Reference U-Net extracts self-attended feature representations from the reference image, which are then used as keys and values in Reference-Attention layers within the corresponding block of the Denoising U-Net~\cite{chen2024echomimic}, as depicted in Figures~\ref{fig:stage1} and~\ref{fig:stage2}. This mechanism enables the model to condition generation on both low-level texture and high-level structural features of the reference face at every spatial scale, ensuring robust identity preservation across all 12 generated frames. Crucially, the Reference U-Net introduces no noise into the reference image and performs only a single forward pass per inference step, incurring minimal additional computational cost.

\subsubsection{Audio Conditioning}

Our initial audio conditioning strategy used Wav2Vec~\cite{schneider2019wav2vec} features, which provide general-purpose speech representations learned through self-supervised training. While these features carry acoustic information, they lack the phoneme-level precision and linguistic alignment that lip motion requires. We subsequently replaced Wav2Vec with Whisper~\cite{radford2023robust}, a large-scale, multilingual speech recognition model trained on 680,000 hours of supervised audio data. Whisper's encoder produces embeddings that are both linguistically structured and robust to acoustic variation, making them significantly better suited for driving fine-grained lip movements. For each generated frame, we construct its audio feature by concatenating the Whisper embeddings of a window of surrounding frames, following the approach of LatentSync~\cite{li2024latentsync}, to provide the model with temporal audio context beyond the instantaneous frame. These audio features are injected into the Denoising U-Net via Audio-Attention cross-attention layers, as shown in Figures~\ref{fig:stage1} and~\ref{fig:stage2}.

\subsection{Data Leakage Analysis and Remediation}

\begin{figure*}[t]
    \centering
    \includegraphics[width=\textwidth]{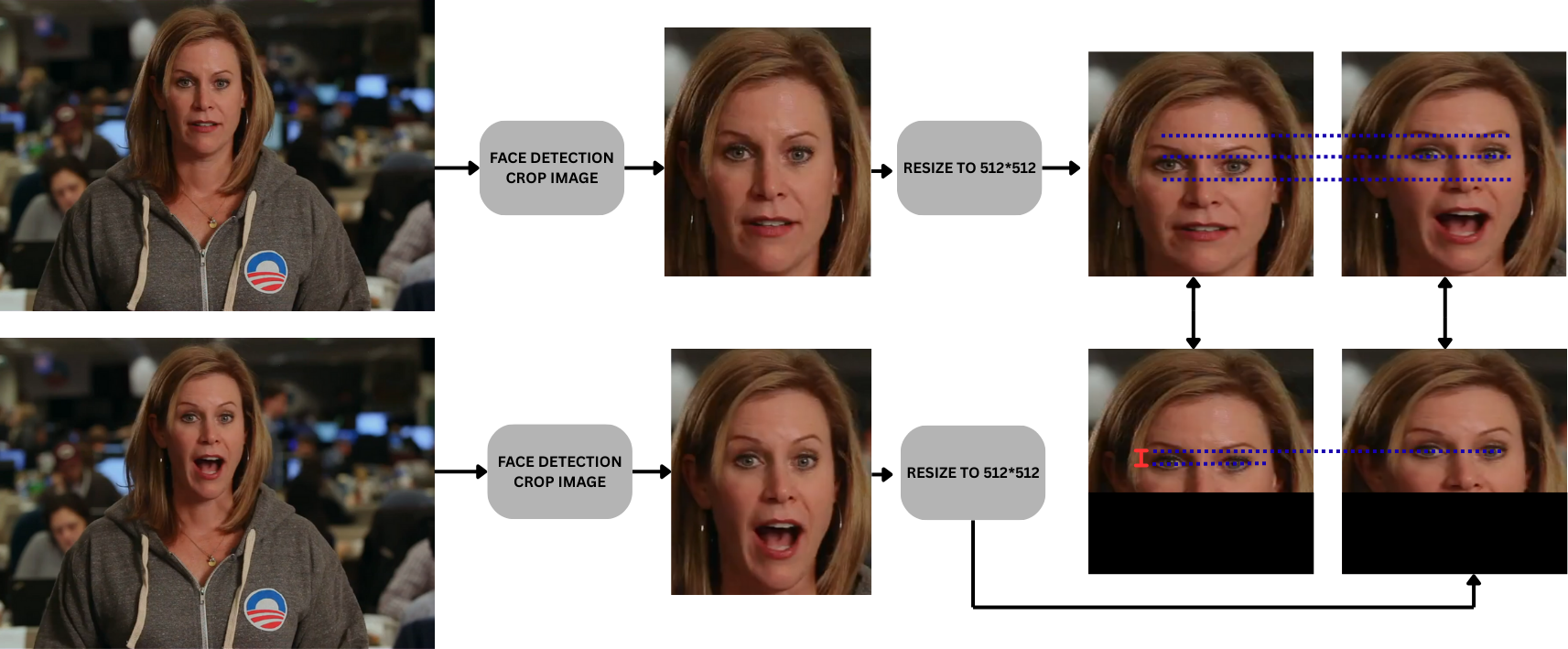}
    \caption{Illustration of data leakage induced by incorrect preprocessing. Per-frame independent face detection produces bounding boxes whose heights are proportional to mouth aperture. After resizing to $512{\times}512$, the upper-face region shifts vertically across frames in proportion to the degree of mouth opening. The blue dashed lines highlight the inconsistent vertical position of the eye region across two frames (mouth closed vs. mouth open), making this positional shift a leakage channel through which the model can infer lip state from the upper face without consulting the audio signal. The bottom row shows the corresponding masked frames fed to the model, in which the lip region is zeroed out but the positional cue in the upper face remains intact.}
    \label{fig:leakage_incorrect}
\end{figure*}

\begin{figure*}[t]
    \centering
    \includegraphics[width=\textwidth]{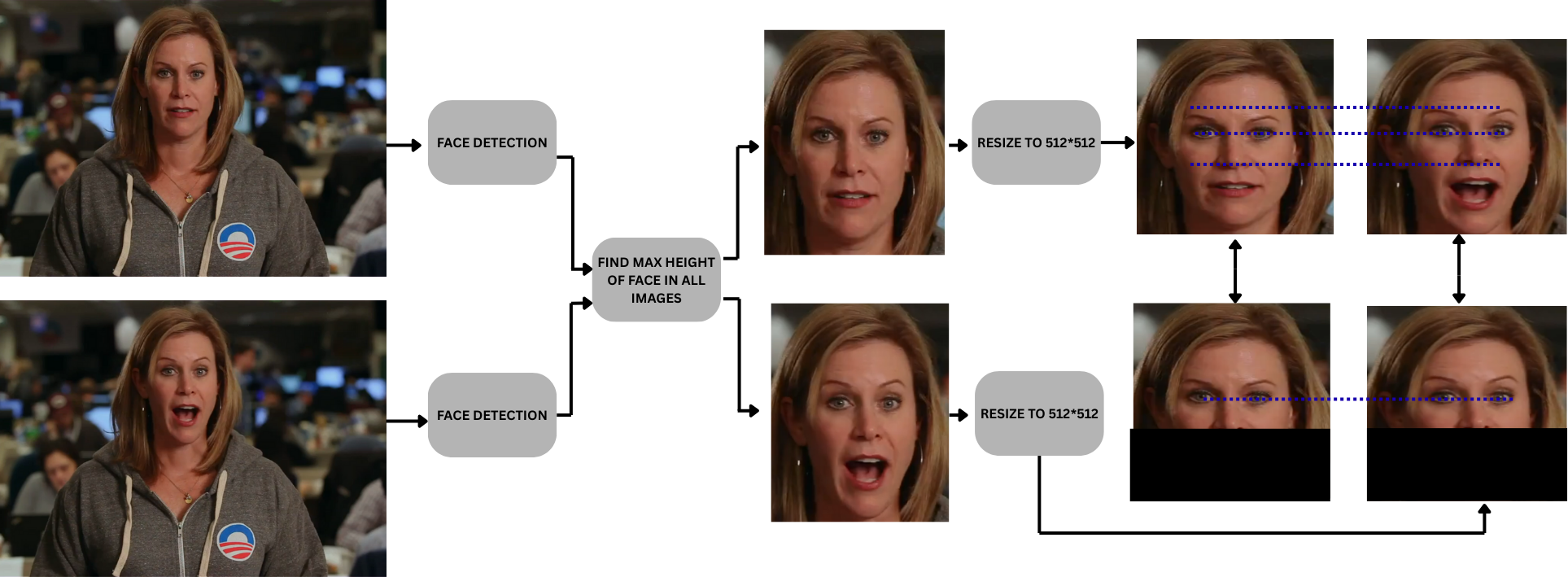}
    \caption{Illustration of the corrected preprocessing pipeline that eliminates height-induced data leakage. The maximum bounding box height across all frames in a video is computed and applied uniformly, so that all crops span the same absolute facial extent. After resizing to $512{\times}512$, the upper-face region appears at a consistent vertical position regardless of mouth aperture, as confirmed by the aligned eye region (blue dashed lines) across frames with different lip states. The model can no longer use upper-face position as a proxy for lip state and must rely on the audio conditioning signal to determine lip shape.}
    \label{fig:leakage_correct}
\end{figure*}

\begin{figure}[t]
    \centering
    \includegraphics[width=\linewidth]{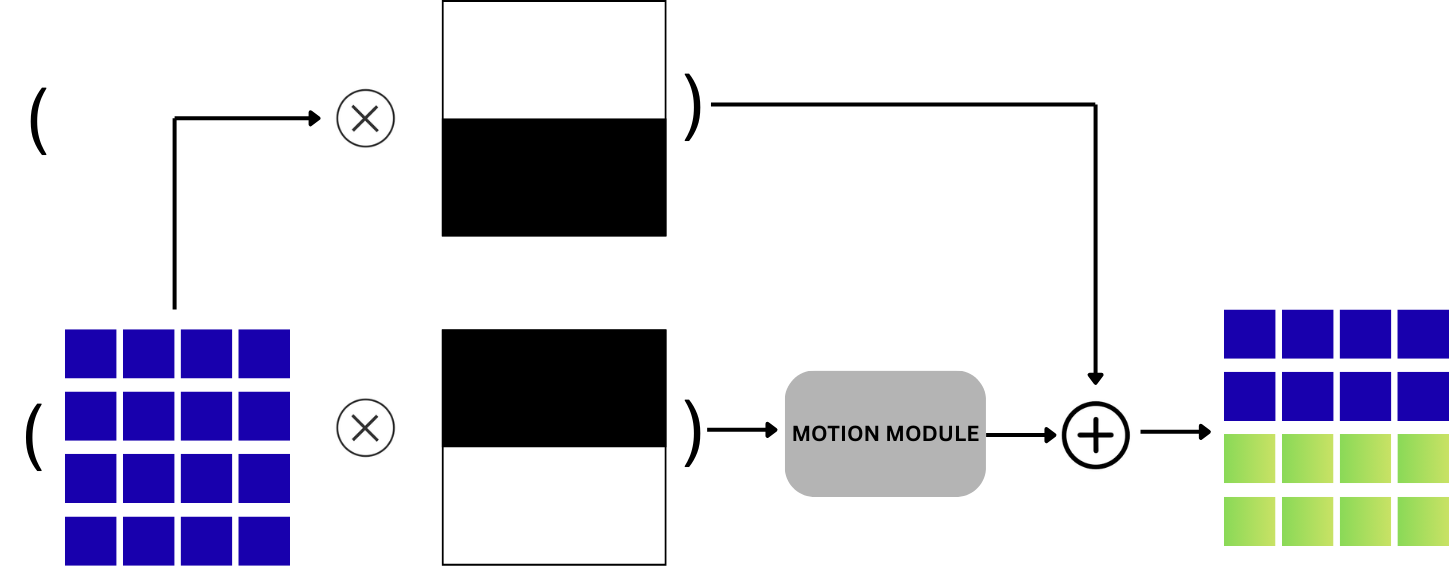}
    \caption{Illustration of the masked attention mechanism applied within the motion module to block upper-face-to-lip leakage. Each frame's feature map (represented as a spatial grid of tokens) is divided into an upper-face region (blue tokens) and a lower-face region (green tokens) using a fixed binary mask. During temporal self-attention, the attention matrix is masked such that lower-face tokens cannot attend to upper-face tokens across the time dimension. The upper-face features are passed through the motion module unmodified via a separate pathway and recombined with the lower-face output via element-wise addition, preserving identity-relevant upper-face information in the final output while preventing it from influencing lip motion synthesis.}
    \label{fig:masked_attn}
\end{figure}

The most consequential finding of this work is the identification and remediation of a systematic data leakage problem that has, in our analysis, been the primary obstacle to effective motion module training in diffusion-based lip sync models. Despite masking the lip region in all input frames to prevent direct exposure of the ground-truth lip state, we consistently observed that the model reproduced the original lip trajectory even when the correct audio was replaced with silence or a mismatched signal. This behavior, which we term \textit{data leakage}, indicates that the model learns to infer lip state from indirect cues present in the unmasked upper-face region, rather than from the audio conditioning signal. The leakage is negligible when frames are processed independently, but becomes strongly amplified when a motion module jointly processes consecutive frames, which is why most prior works that process frames independently~\cite{prajwal2020lip, mukhopadhyay2024diff2lip, li2024latentsync} have not reported it. We identify two mechanistically distinct sources, illustrated in Figures~\ref{fig:leakage_incorrect},~\ref{fig:leakage_correct}, and~\ref{fig:masked_attn}.

\textbf{Leakage Source 1: Face height variation from per-frame preprocessing.} Standard face detection pipelines localize and crop the face region independently in each frame, producing a bounding box that extends from the top of the head to the base of the jaw. As a direct consequence, the bounding box height varies with mouth aperture: it is larger when the mouth is open and smaller when the mouth is closed. After resizing all crops to a fixed $512{\times}512$ resolution, this height variation manifests as a relative shift in the vertical position of the upper face (eyes, nose) across frames~\cite{zhang2024musetalk}, as illustrated in Figure~\ref{fig:leakage_incorrect}. When the model processes 12 such consecutive frames jointly, this systematic co-variation between upper-face position and lip state provides a strong implicit cue that the model exploits to recover the original lip trajectory---without any need to attend to the audio signal.

We eliminate this leakage source by computing the maximum face bounding box height across all frames in a given video and using this fixed height uniformly for all crops, as depicted in Figure~\ref{fig:leakage_correct}. This normalization removes frame-to-frame height variation as an informative signal, ensuring that the vertical position of the upper face is constant regardless of lip state.

\textbf{Leakage Source 2: Biomechanical coupling between upper and lower facial muscles.} Beyond the preprocessing artifact described above, a subtler leakage channel exists due to the anatomical connectivity between the muscles of the upper and lower face~\cite{li2024latentsync}. Although this coupling is weak and would likely be ignored by a model processing frames independently, it becomes a tractable signal when exploited across time by a motion module performing temporal self-attention. The motion module can learn to correlate upper-face dynamics across consecutive frames with the corresponding lip movements, effectively sidestepping the audio conditioning.

We address this by introducing a \textit{spatially masked attention mechanism} within the motion module, illustrated in Figure~\ref{fig:masked_attn}. Specifically, the spatial feature maps of each frame are partitioned into an upper-face region and a lower-face (lip) region using a binary mask. During temporal self-attention inside the motion module, attention weights from lower-face tokens to upper-face tokens are masked to zero, severing the pathway through which upper-face dynamics could inform lip state reconstruction. This forces the model to derive all temporal lip motion information from the audio signal and from the lower-face tokens themselves.

A third, more subtle leakage source relates to mask image serialization. Saving binary segmentation masks in JPEG format---a lossy compression codec---introduces non-zero residuals within the nominally zeroed lip region due to block-based quantization artifacts. These residual pixel values partially expose the original lip texture through the mask, constituting an additional leakage pathway. We enforce lossless PNG serialization for all mask images throughout training to eliminate this effect entirely.

\subsection{High-Resolution Generation at 512$\times$512}

Achieving lip sync at $512{\times}512$ resolution introduces non-trivial computational and modeling challenges. To manage these while preserving visual fidelity, we employ the VAE encoder-decoder pipeline of SD 1.5~\cite{rombach2022high} to compress all input and output frames into a $64{\times}64$ latent representation, reducing the effective spatial dimensionality by a factor of 64 while retaining perceptually relevant structure. This approach follows the standard latent diffusion paradigm~\cite{rombach2022high} and significantly reduces both training memory requirements and inference latency compared to pixel-space diffusion~\cite{mukhopadhyay2024diff2lip}.

At this resolution, training a SyncNet discriminator is particularly problematic: prior work has identified widespread SyncNet convergence failures even at $256{\times}256$~\cite{li2024latentsync}, and the difficulty is compounded at higher resolutions. Rather than relying on this fragile external supervision, we discard SyncNet entirely and supervise synchronization implicitly through a simple pixel-space reconstruction loss computed after VAE decoding:
\begin{equation}
    \mathcal{L}_{\text{spatial}} = \|\hat{\mathbf{x}} - \mathbf{x}\|_2
\end{equation}
where $\hat{\mathbf{x}}$ and $\mathbf{x}$ denote the decoded predicted frame and the ground-truth frame, respectively. This deceptively simple objective proves highly effective when combined with our motion module: because the motion module enforces temporal consistency across 12 frames, the reconstruction loss implicitly penalizes temporally incoherent lip trajectories and drives the model toward audio-consistent motion without requiring adversarial supervision.

\subsection{Asymmetric Reference Strategy}

During Stage 2 training, we deliberately use a single randomly selected reference image shared across all 12 frames in each training batch. This design choice prevents the model from exploiting a trivial shortcut, copying lip texture directly from a frame-matched reference, which would suppress the model's need to generate lip motion from the audio signal. However, during inference, we provide each of the 12 frames with its own corresponding ground-truth frame as an individual reference. This asymmetric strategy gives the model access to precise per-frame identity and texture information at test time, substantially reducing generation artifacts without compromising the audio-driven synchronization behavior learned during training.

\subsection{Temporal Coherence and Long-Form Video Generation}

\subsubsection{Motion Module}

The integration of a temporal motion module, successfully achieved in HighSync for the first time in the diffusion-based lip sync literature, in contrast to the reported failure of LatentSync~\cite{li2024latentsync}, is the direct beneficiary of our data leakage remediation. The motion module applies self-attention across the temporal dimension of the 12-frame sequence within each transformer block of the Denoising U-Net, following the AnimateDiff-style temporal layer design~\cite{guo2023animatediff}. By attending jointly to all frames in the sequence, it learns to produce smooth, physically plausible lip trajectories rather than independently generated per-frame outputs. As we demonstrate empirically, this temporal modeling capability is only achievable once data leakage is eliminated: in its presence, the motion module converges to exploit leakage cues rather than audio features, producing accurate-looking but audio-independent lip motion.

\subsubsection{Cross-Group Temporal Continuity}

Generating videos longer than 12 frames requires a mechanism for maintaining coherence across adjacent generation groups. We evaluated two candidate strategies. The first, inspired by Hallo~\cite{xu2024hallo}, bridges consecutive groups by passing the last two generated frames of the current group as explicit motion conditioning for the next. In practice, this approach introduced a progressive accumulation of noise with each successive group, making it unsuitable for long-form generation. The second strategy, adapted from EchoMimic~\cite{chen2024echomimic}, employs overlapped diffusion context: the denoising trajectories of the boundary frames of one group are shared with the initial frames of the next group across all diffusion timesteps. This soft temporal coupling proved both stable and effective, and constitutes our adopted solution for cross-group continuity.

\subsubsection{Memory-Efficient Streaming Generation}

The naive application of overlapped diffusion context requires maintaining the full denoising trajectory of all video frames simultaneously in GPU memory, which becomes infeasible for videos of non-trivial duration. We resolve this by implementing a streaming generation scheme: the intermediate diffusion states of the last two frames of each generation round are cached and reused as overlapping context for the subsequent round, while all other frames are discarded. This allows HighSync to generate long-form video while requiring only  10~GB of GPU memory, with seamless temporal continuity maintained across rounds via the cached boundary states.

\subsection{Two-Stage Training Procedure}
We adopt a two-stage training strategy, depicted in Figures~\ref{fig:stage1} and~\ref{fig:stage2}, that progressively decouples visual quality learning from temporal motion learning, similar to EchoMimic~\cite{chen2024echomimic, xu2024hallo}:

\begin{itemize}
    \item \textbf{Stage 1} (Figure~\ref{fig:stage1}): The full model---excluding the motion module---is trained end-to-end on single-frame data. This stage establishes robust visual generation quality, identity preservation via the Reference U-Net, and foundational audio conditioning via Whisper cross-attention, in the absence of any temporal modeling pressure that could induce condition dominance.
    \item \textbf{Stage 2} (Figure~\ref{fig:stage2}): All Stage 1 components are frozen and only the motion module is trained on 12-frame video sequences. Freezing the visual and audio pathways ensures that the motion module learns to produce temporally coherent lip motion in response to the already-learned audio features, rather than relearning the conditioning signals from scratch under potentially destabilizing temporal dynamics.
\end{itemize}

Throughout both stages, we apply classifier-free guidance dropout~\cite{ho2021classifier} at a rate of 0.1 exclusively to the reference image condition, randomly zeroing out reference features to prevent the model from exclusively relying on visual identity. We deliberately omit dropout from the audio condition: given that audio is inherently the weaker conditioning signal, further attenuation during training would risk its complete suppression and undermine synchronization learning.

\section{Experiments}

\subsection{Dataset Curation}

A prerequisite for training a $512{\times}512$ lip sync model is access to high-resolution video datasets in which the cropped face region natively exceeds the target resolution. Standard benchmark datasets widely used in prior lip sync work, including LRS2~\cite{afouras2018deep}, LRS3~\cite{afouras2018lrs3}, and LRW~\cite{chung2016lip}, are captured at insufficient resolution for our purposes and cannot be upsampled without introducing the very artifacts we seek to avoid. We construct our training corpus from three high-quality video datasets: VFHQ~\cite{xie2022vfhq}, HDTF~\cite{zhang2021flow}, and CelebV-HQ~\cite{zhu2022celebvhq}.

\textbf{VFHQ}~\cite{xie2022vfhq} is a large-scale, multilingual face video dataset curated for super-resolution research, providing high native resolution across diverse identities and recording conditions. Of its 16,000 videos, 7,500 passed our quality filtering criteria.

\textbf{HDTF}~\cite{zhang2021flow} is an English-language dataset of high-definition talking-head recordings, characterized by predominantly frontal pose, controlled backgrounds, and stable camera conditions. Its 362 long-form videos were segmented into 10-second clips, yielding approximately 7,000 segments, of which 6,000 were retained after cleaning.

\textbf{CelebV-HQ}~\cite{zhu2022celebvhq} is a large-scale, in-the-wild multilingual dataset spanning a wide range of identities, head poses, lighting conditions, and occlusion scenarios---exactly the challenging conditions for which our model is designed. Of its 35,000 videos, 7,000 were retained after filtering.

All videos were downloaded at maximum available quality. Automated filtering removed clips with insufficient face resolution, excessive motion blur, scene transitions, background changes, and multi-speaker segments. Given the limitations of automated tools in detecting subtler quality issues, such as inherently weak audio-lip synchronization or intermittent but non-persistent facial occlusions, we conducted an additional manual cleaning phase involving five trained annotators over six days. Table~\ref{tab:dataset} summarizes the dataset statistics before and after each cleaning stage, underscoring the necessity of manual curation for constructing a training corpus of sufficient quality. The curated datasets are publicly released to facilitate reproducibility and community use.\footnote{
    \url{https://huggingface.co/datasets/saeed-5959/vfhq} \\
    \url{https://huggingface.co/datasets/saeed-5959/celebv_hq_head_talking} \\
    \url{https://huggingface.co/datasets/saeed-5959/hdtf}
}

\begin{table}[t]
\centering
\caption{Dataset statistics before and after automated and manual cleaning.}
\label{tab:dataset}
\begin{tabular}{lccc}
\toprule
\textbf{Dataset} & \textbf{Original} & \textbf{After Auto} & \textbf{After Manual} \\
\midrule
VFHQ      & 16,000 & 10,000 & 7,500 \\
HDTF      & 7,000 & 6,500 & 6,000 \\
CelebV-HQ & 35,000 & 20,000 & 7,000 \\
\bottomrule
\end{tabular}
\end{table}

\subsection{Implementation Details}

All experiments were conducted on a single NVIDIA A100 GPU with 80~GB of memory. Input videos were  first processed by detecting and cropping the face region, then resampled to 25~FPS and resized to $512{\times}512$ pixels; audio was resampled to 16~kHz prior to Whisper feature extraction. Stage 1 training used the AdamW optimizer with a learning rate of $1{\times}10^{-5}$ and a batch size of 16 for 300,000 steps. Stage 2 training used the same optimizer with a batch size of 4---each element comprising 12 consecutive frames---for 100,000 steps. Reference image dropout was applied at a rate of 0.1 in both stages. Inference was performed using 20 DDIM~\cite{song2020denoising} sampling steps.

\begin{table*}[t]
\centering
\caption{Quantitative comparison of HighSync against state-of-the-art methods across three benchmark datasets. Best results among generated methods are highlighted in \textbf{bold}.}
\label{tab:quantitative}
\setlength{\tabcolsep}{4pt}
\begin{tabular}{lcccccccccc}
\toprule
& \multicolumn{3}{c}{\textbf{VFHQ}} & \multicolumn{3}{c}{\textbf{HDTF}} & \multicolumn{3}{c}{\textbf{CelebV-HQ}} \\
\cmidrule(lr){2-4}\cmidrule(lr){5-7}\cmidrule(lr){8-10}
\textbf{Method} & FID$\downarrow$ & CSIM$\uparrow$ & LSE-C$\uparrow$ & FID$\downarrow$ & CSIM$\uparrow$ & LSE-C$\uparrow$ & FID$\downarrow$ & CSIM$\uparrow$ & LSE-C$\uparrow$ \\
\midrule
Wav2Lip~\cite{prajwal2020lip}        & 14.85 & 0.82 & 6.58 & 12.25 & 0.82 & 7.38 & 16.18 & 0.78 & 6.43 \\
VideoReTalking~\cite{cheng2022videoretlaking}  & 14.35 & 0.80 & 6.45 & 12.04 & 0.81 & 7.01 & 15.68 & 0.79 & 6.35 \\
Diff2Lip~\cite{mukhopadhyay2024diff2lip}       & 16.74 & 0.78 & 6.24 & 13.45 & 0.79 & 7.08 & 16.98 & 0.74 & 6.12 \\
LatentSync~\cite{li2024latentsync}     &  9.78 & 0.85 & 6.75 &  8.98 & 0.84 & 7.58 & 10.12 & 0.83 & 6.51 \\
MuseTalk~\cite{zhang2024musetalk}       &  7.49 & 0.85 & 5.25 &  \textbf{7.25} & \textbf{0.86} & 6.21 &  8.37 & \textbf{0.84} & 4.85 \\
\midrule
Ground Truth    & 0.00 & 1.00 & 7.11 & 0.00 & 1.00 & 7.78 & 0.00 & 1.00 & 6.97 \\
\midrule
\textbf{HighSync (Ours)} & \textbf{7.22} & \textbf{0.86} & \textbf{7.02} & {7.36} & {0.85} & \textbf{7.72} & \textbf{8.15} & \textbf{0.84} & \textbf{6.75} \\
\bottomrule
\end{tabular}
\end{table*}

\subsection{Evaluation Protocol}
 
We evaluate HighSync across three complementary dimensions.
 
\textbf{Visual quality} is assessed using the Fr\'echet Inception Distance (FID)~\cite{heusel2017gans}, which measures the distributional similarity between generated and real image sets, with lower values indicating greater realism, and the Cosine Similarity of Identity embeddings ~\cite{deng2019arcface} (CSIM), which quantifies how faithfully the generated frames preserve the identity of the source subject, with higher values indicating stronger identity consistency.

\textbf{Lip-sync accuracy} is measured using the Lip Sync Error Confidence score (LSE-C)~\cite{prajwal2020lip}, which reflects the degree of audio-visual alignment as assessed by an independent discriminator, with higher values indicating tighter synchronization.
 
All three metrics are evaluated on three datasets, VFHQ~\cite{xie2022vfhq}, HDTF~\cite{zhang2021flow}, and CelebV-HQ~\cite{zhu2022celebvhq}, and results are reported in Table~\ref{tab:quantitative}.
 
Beyond these standard metrics, we introduce the \textit{silence test} as a novel diagnostic evaluation specifically designed to probe the degree to which a model's output is genuinely conditioned on the audio signal, rather than on visual leakage cues. In this evaluation, a silent audio clip of five seconds duration, containing no speech signal whatsoever, is provided as input alongside a source video. A model that has learned true audio conditioning will respond to the absence of a speech signal by generating closed-lip output throughout the sequence. Conversely, a model that relies on visual leakage or has failed to learn meaningful audio dependence will replicate the original lip motion regardless of the silent input. The metric is defined as the fraction of generated frames in which the lips are assessed to be in a closed state, over the total number of frames in the sequence. A score approaching 1.0 indicates strong audio conditioning; a score near 0.0 indicates that the model largely ignores the audio. Results for all compared methods are reported in Table~\ref{tab:silence}.
 
Recognizing that automated metrics capture only a subset of the perceptual dimensions relevant to real-world deployment, we additionally conduct a structured human evaluation study, similar to Wav2Lip ~\cite{prajwal2020lip} and MuseTalk ~\cite{zhang2024musetalk}. Fifty video clips of 20 to 60 seconds in duration were selected from sources outside our three training datasets. Ten participants independently rated the output of each method on two axes: (1) overall image quality and (2) lip-synchronization quality, using a five-point scale (1~=~worst, 5~=~best). To ensure unbiased assessment, method names were withheld from all participants and all videos were presented in randomized order, such that no participant could associate a rating with a specific method. A total of 500 ratings per method were collected across both axes. Results are presented in Table~\ref{tab:human}.
 
\subsection{Quantitative Evaluation}
 
\textbf{Comparison with state-of-the-art methods.}
Table~\ref{tab:quantitative} presents the quantitative comparison of HighSync against five competing methods, Wav2Lip~\cite{prajwal2020lip}, VideoReTalking~\cite{cheng2022videoretlaking}, Diff2Lip~\cite{mukhopadhyay2024diff2lip}, LatentSync~\cite{li2024latentsync}, and MuseTalk~\cite{zhang2024musetalk}, across all three evaluation datasets. HighSync achieves the best or second-best FID score across all three datasets, demonstrating that its $512{\times}512$ diffusion-based generation produces frame-level visual quality that closely approximates the real image distribution. Specifically, HighSync achieves an FID of 7.22 on VFHQ, 7.36 on HDTF, and 8.15 on CelebV-HQ, consistently outperforming GAN-based methods such as Wav2Lip~\cite{prajwal2020lip} and VideoReTalking~\cite{cheng2022videoretlaking} by a substantial margin, and matching or surpassing the diffusion-based LatentSync~\cite{li2024latentsync} despite operating at twice the resolution. In terms of identity preservation (CSIM), HighSync achieves 0.86, 0.85, and 0.84 across the three datasets, reflecting strong fidelity to the source subject's appearance, a direct consequence of the Reference U-Net conditioning mechanism. On the lip-sync accuracy metric (LSE-C), HighSync achieves scores of 7.02, 7.72, and 6.75 on VFHQ, HDTF, and CelebV-HQ respectively, surpassing all diffusion-based and GAN-based methods, notably without relying on SyncNet supervision during training. It is worth emphasizing that these synchronization results are achieved purely through motion module training with a simple spatial reconstruction loss, without any adversarial synchronization discriminator, which underscores the effectiveness of our data-leakage-free temporal modeling approach.

\textbf{Silence test.}
Table~\ref{tab:silence} reports the silence test scores across all compared methods. HighSync achieves a score of 0.93, substantially outperforming all baselines. GAN-based methods, Wav2Lip~\cite{prajwal2020lip} (0.84), VideoReTalking~\cite{cheng2022videoretlaking} (0.82), and MuseTalk~\cite{zhang2024musetalk} (0.68), and diffusion-based methods, Diff2Lip~\cite{mukhopadhyay2024diff2lip} (0.78) and LatentSync~\cite{li2024latentsync} (0.81), all produce notably lower silence scores, indicating that a significant fraction of their generated frames exhibit open-lip motion even in the absence of any speech signal. This behavior is consistent with the data leakage hypothesis: these models learn to partially reconstruct original lip trajectories from visual context rather than from the audio signal, and therefore cannot reliably suppress lip motion when presented with silence. The markedly higher score of HighSync directly validates the effectiveness of our leakage remediation strategy, both the bounding-box height normalization and the masked attention mechanism, in forcing the model to depend genuinely on the audio conditioning signal for all lip motion decisions.
 
\begin{table}[t]
\centering
\caption{Silence test results. The SILENT score~($\uparrow$) denotes the fraction of frames with closed lips when a silent audio input is provided. Higher scores indicate stronger and more genuine audio conditioning.}
\label{tab:silence}
\begin{tabular}{lc}
\toprule
\textbf{Method} & \textbf{SILENT}~$\uparrow$ \\
\midrule
Wav2Lip~\cite{prajwal2020lip}          & 0.84 \\
VideoReTalking~\cite{cheng2022videoretlaking}    & 0.82 \\
Diff2Lip~\cite{mukhopadhyay2024diff2lip}         & 0.78 \\
LatentSync~\cite{li2024latentsync}       & 0.81 \\
MuseTalk~\cite{zhang2024musetalk}         & 0.68 \\
\textbf{HighSync (Ours)}  & \textbf{0.93} \\
\bottomrule
\end{tabular}
\end{table}
 
\textbf{Human evaluation.}
Table~\ref{tab:human} presents the results of the human judgment study. HighSync achieves mean scores of 4.28 for image quality and 4.01 for synchronization quality, the highest synchronization score among all compared methods and the second-highest image quality score after MuseTalk~\cite{zhang2024musetalk} (4.34 quality, 3.14 sync). This result highlights an important trade-off present in prior work: MuseTalk achieves high perceptual image quality through its GAN-based adversarial refinement, but participants rated its synchronization quality substantially lower than HighSync (3.14 vs. 4.01), suggesting that its one-step generation without genuine temporal audio conditioning produces visually appealing but poorly synchronized lip motion. LatentSync~\cite{li2024latentsync} achieves the second-highest synchronization score (3.68) among baselines, consistent with its strong automated metrics, but lags behind HighSync by a considerable margin. GAN-based methods Wav2Lip~\cite{prajwal2020lip} and VideoReTalking~\cite{cheng2022videoretlaking} receive the lowest image quality scores (3.78 and 3.54, respectively), reflecting the well-known visual degradation associated with their low-resolution training regimes. Diff2Lip~\cite{mukhopadhyay2024diff2lip} receives the lowest image quality score overall (2.15), indicating that pixel-space diffusion at low resolution introduces perceptible artifacts that human observers readily identify. The Ground Truth achieves scores of 4.78 and 4.35, providing a perceptual upper bound and confirming that HighSync's outputs are competitive with real video, with only a modest gap remaining in both dimensions. All participants provided informed consent prior to participation, 
and no personally identifiable information was collected.
 
\begin{table}[t]
\centering
\caption{Human judgment evaluation results. Participants rated generated videos on a 1--5 scale across image quality (QUALITY~$\uparrow$) and lip-synchronization quality (SYNC~$\uparrow$). Scores represent the mean across all participants and clips. Ground Truth ratings provide a perceptual upper bound.}
\label{tab:human}
\begin{tabular}{lcc}
\toprule
\textbf{Method} & \textbf{QUALITY}~$\uparrow$ & \textbf{SYNC}~$\uparrow$ \\
\midrule
Ground Truth                              & 4.78 & 4.35 \\
\midrule
Wav2Lip~\cite{prajwal2020lip}             & 3.78 & 3.22   \\
VideoReTalking~\cite{cheng2022videoretlaking} & 3.54 & 3.02   \\
Diff2Lip~\cite{mukhopadhyay2024diff2lip}  & 2.15 & 3.15   \\
LatentSync~\cite{li2024latentsync}        & 3.95   & 3.68 \\
MuseTalk~\cite{zhang2024musetalk}         & \textbf{4.34} & 3.14 \\
\textbf{HighSync (Ours)}                  & {4.28} & \textbf{4.01} \\
\bottomrule
\end{tabular}
\end{table}

\subsection{Qualitative Evaluation}

\begin{figure*}[!t]
    \centering
    \includegraphics[width=\textwidth]{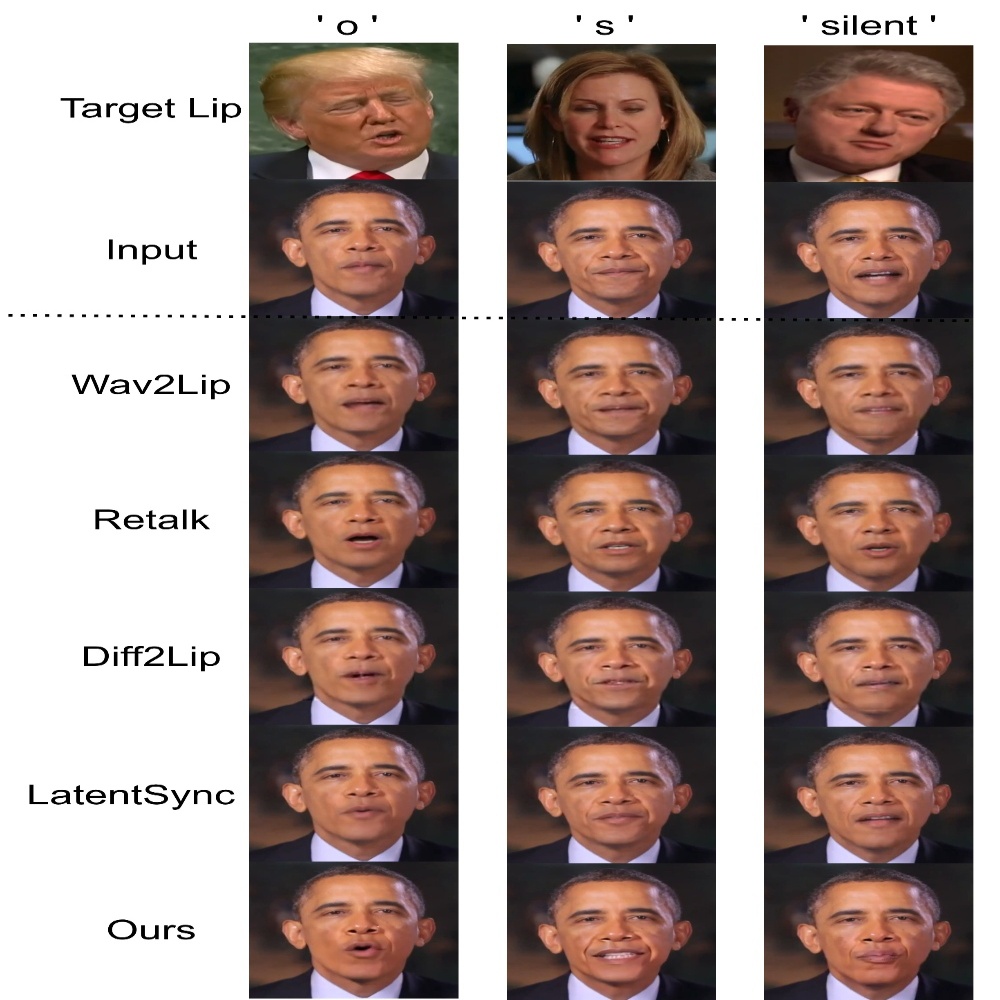}
    \caption{Qualitative comparison of lip synchronization results across six methods for three audio conditions: the vowel /o/, the fricative /s/, and silence. For each condition, the target lip shape (derived from the audio source speaker) is shown in the top row, followed by the input identity frame and the outputs of Wav2Lip~\cite{prajwal2020lip}, VideoReTalking~\cite{cheng2022videoretlaking}, Diff2Lip~\cite{mukhopadhyay2024diff2lip}, LatentSync~\cite{li2024latentsync}, and HighSync (Ours). HighSync consistently produces the most accurate phoneme-to-viseme correspondences, a rounded aperture for /o/, a partially open, teeth-visible shape for /s/ and complete lip closure for silence, while maintaining high visual fidelity and identity consistency throughout.}
    \label{fig:qualitative}
\end{figure*}

Figure~\ref{fig:qualitative} presents a qualitative comparison of all methods across three phonetically distinct audio conditions, the vowel sound /o/, the fricative /s/ and silence, using a fixed source identity driven by audio from three different target speakers. The figure illustrates that HighSync produces lip shapes that closely match the phonetic content of the driving audio: a rounded, open-mouth shape for /o/, a narrow, teeth-visible configuration for /s/, and a fully closed mouth for silence. Competing methods show varying degrees of failure: Wav2Lip~\cite{prajwal2020lip} and VideoReTalking~\cite{cheng2022videoretlaking} produce blurry lip regions with imprecise phoneme-to-viseme correspondence; Diff2Lip~\cite{mukhopadhyay2024diff2lip} exhibits incorrect lip shapes and visible artifacts in the teeth region, with notably poor synchronization under the /o/ vowel condition; LatentSync~\cite{li2024latentsync} demonstrates reasonable synchronization but inconsistent closure under the silence condition, confirming the leakage behavior reflected in Table~\ref{tab:silence}.

\begin{figure*}[!t]
    \centering
    \includegraphics[width=\textwidth]{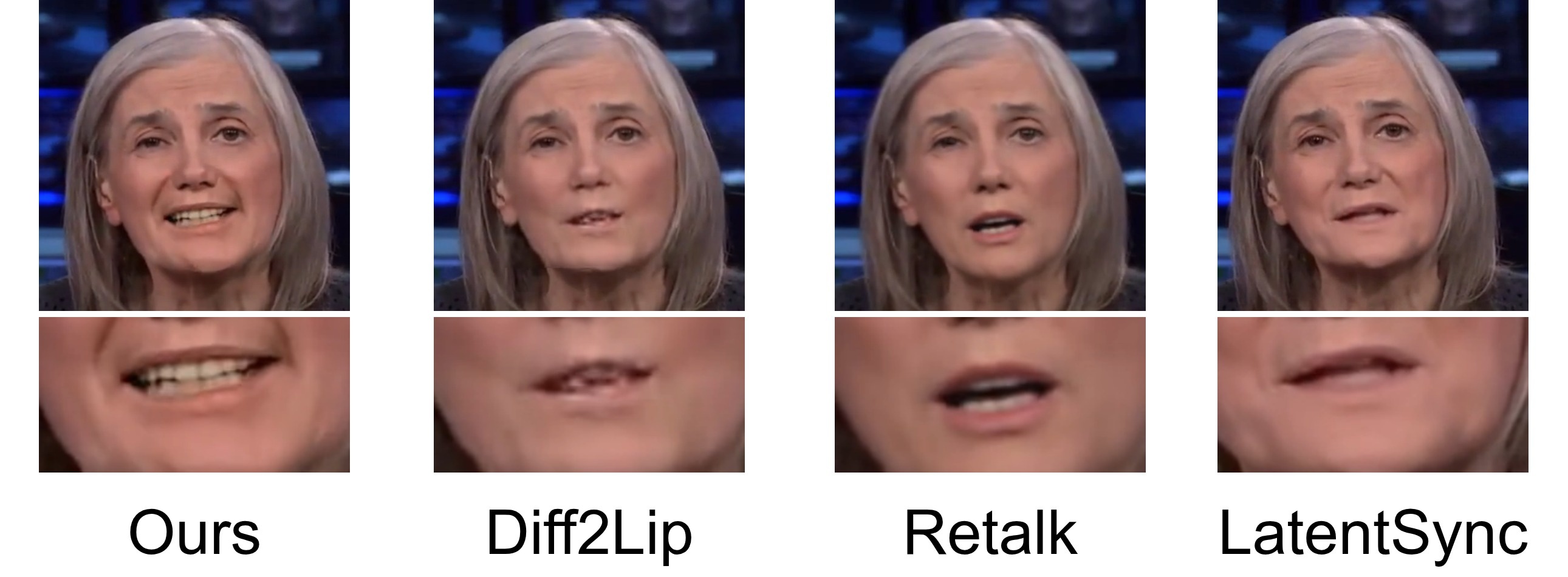}
    \caption{Qualitative comparison of generated lip and teeth quality across four methods. The top row shows the full face output; the bottom row shows a zoomed crop of the lip and teeth region. HighSync (Ours) produces the most anatomically detailed and visually realistic teeth and lip textures, with clearly defined tooth boundaries. Diff2Lip~\cite{mukhopadhyay2024diff2lip} produces an unnaturally dark interior with imprecise lip boundaries. VideoReTalking~\cite{cheng2022videoretlaking} generates a recognizable but blurred dental region with loss of fine structure. LatentSync~\cite{li2024latentsync} generates a smooth but structurally ambiguous mouth region, lacking the fine-grained dental detail visible in HighSync's outputs.}
    \label{fig:teeth}
\end{figure*}

Figure~\ref{fig:teeth} provides a close-up qualitative comparison of the lip and teeth region across HighSync, Diff2Lip~\cite{mukhopadhyay2024diff2lip}, VideoReTalking~\cite{cheng2022videoretlaking}, and LatentSync~\cite{li2024latentsync}. The figure demonstrates a clear advantage of HighSync in terms of dental detail and lip texture: our model generates anatomically plausible teeth structures with realistic gum boundaries and individual tooth definition, whereas competing methods produce smeared, undifferentiated, or implausible dental regions. This improvement can be attributed to our $512{\times}512$ latent diffusion architecture, which preserves high-frequency detail in the VAE latent space, and to the Spatial Loss applied directly in pixel space, which explicitly penalizes perceptual degradation in the generated region.

\subsection{Ablation Study: Data Leakage Remediation}

To isolate the contribution of each proposed leakage remediation strategy, we conduct an ablation study on the VFHQ dataset, evaluating four configurations of HighSync: the baseline without either intervention, with normalized preprocessing only, with masked attention only, and with both components combined. Results are reported in Table~\ref{tab:ablation}.

The results reveal a clear and consistent pattern. The baseline configuration, without either intervention, achieves an LSE-C of only 3.15, confirming that in the presence of data leakage the motion module exploits visual cues rather than the audio signal, producing severely degraded synchronization despite reasonable visual quality (FID 7.23, CSIM 0.86). Introducing normalized preprocessing alone raises LSE-C substantially to 6.12, demonstrating that eliminating bounding-box height variation is the more impactful of the two interventions. Adding masked attention alone yields an LSE-C of 5.24, confirming that suppressing upper-face-to-lip attention pathways provides an independent and complementary synchronization benefit. When both strategies are applied jointly, LSE-C reaches 7.02, the best result across all configurations, while FID and CSIM remain competitive, confirming that the two remediation strategies address distinct leakage channels and that their combination is necessary to fully eliminate the phenomenon.

\begin{table*}[t]
\centering
\caption{Ablation study on the VFHQ dataset evaluating the individual and combined contributions of the two data leakage remediation strategies. FID~($\downarrow$), CSIM~($\uparrow$), LSE-C~($\uparrow$).}
\label{tab:ablation}
\begin{tabular}{lccc}
\toprule
\textbf{Configuration} & \textbf{FID}$\downarrow$ & \textbf{CSIM}$\uparrow$ & \textbf{LSE-C}$\uparrow$ \\
\midrule
HighSync w/o normalized preprocess and w/o masked attention & 7.23 & 0.86 & 3.15 \\
HighSync + normalized preprocess                            & 7.19 & 0.85 & 6.12 \\
HighSync + masked attention                                & 7.18 & 0.84 & 5.24 \\
HighSync + normalized preprocess + masked attention         & \textbf{7.22} & \textbf{0.86} & \textbf{7.02} \\
\bottomrule
\end{tabular}
\end{table*}

\section{Conclusion}
 
We have presented HighSync, an end-to-end latent diffusion framework for lip synchronization that simultaneously advances visual generation quality and audio-driven synchronization accuracy. By extending Stable Diffusion 1.5 with a dedicated Reference U-Net, Whisper-based audio cross-attention, and a temporally-aware motion module, HighSync generates photorealistic, temporally coherent talking-face videos at $512{\times}512$ resolution, a resolution barrier no prior lip sync model has achieved.

Our contributions to this work consist of two major components. The first is the construction of an end-to-end high-quality lip sync pipeline operating at $512{\times}512$ resolution, enabled by latent diffusion modeling and a two-stage training strategy that decouples visual quality learning from temporal motion modeling. The second, and more fundamental, contribution is the identification and elimination of the data leakage problem that has silently prevented prior models from achieving genuine audio conditioning under temporal modeling. We demonstrate that leakage arises from two independent sources: Per-frame bounding box height variation and the biomechanical correlation between upper facial dynamics and lip movement, and propose concrete remediation strategies for both. The effectiveness of these interventions is directly validated by our silence test, in which HighSync achieves a score of 0.93, substantially outperforming all baselines.

Comprehensive quantitative and human evaluations demonstrate that HighSync achieves the best overall balance between visual fidelity, identity preservation, and synchronization accuracy among all compared methods, with human ratings of 4.01 for synchronization and 4.28 for image quality, closely approaching ground-truth performance. We release all model weights, training code, and curated datasets to support reproducibility and future research.

\bibliographystyle{IEEEtran}

\end{document}